\documentclass[sigconf]{acmart}
\usepackage{graphicx} 
\usepackage{subcaption} 
\newtheorem{problem}{Problem}
\usepackage{caption}
\usepackage{float} 
\usepackage{multirow}

\usepackage{algorithm}
\usepackage[noend]{algpseudocode}

\AtBeginDocument{%
  }

\copyrightyear{2025} 
\acmYear{2025} 
\setcopyright{acmlicensed}\acmConference[CIKM '25]{Proceedings of the 34th ACM International Conference on Information and Knowledge
Management}{November 10--14, 2025}{Seoul, Republic of Korea}
\acmBooktitle{Proceedings of the 34th ACM International Conference on
Information and Knowledge Management (CIKM '25), November 10--14, 2025,
Seoul, Republic of Korea}
\acmDOI{10.1145/3746252.3761134}
\acmISBN{979-8-4007-2040-6/2025/11}

\begin{document}

\title{EvoFormer: Learning Dynamic Graph-Level Representations with Structural and Temporal Bias Correction }

\author{Haodi Zhong}
\orcid{0000-0001-6033-4959}
\affiliation{%
  \institution{Xidian University}
    \city{Xian}
    \country{China}  
}
\email{zhonghaodi@xidian.edu.cn}

\author{Liuxin Zou}
\orcid{0009-0004-5500-0910}
\affiliation{%
  \institution{Xidian University}
      \city{Xian}
    \country{China} 
}
\email{23031212251@stu.xidian.edu.cn}

\author{Di Wang}
\orcid{0000-0001-8027-4287}
\affiliation{%
   \institution{Xidian University}
       \city{Xian}
    \country{China} 
}
\email{wangdi@xidian.edu.cn}
\authornote{Corresponding author.}

\author{Bo Wan}
\orcid{0000-0002-3410-9560}
\affiliation{%
 \institution{Xidian University}
     \city{Xian}
    \country{China} 
 }
 \email{wanbo@xidian.edu.cn}

\author{Zhenxing Niu}
\orcid{0000-0003-1660-5138}
\affiliation{%
  \institution{Xidian University}
      \city{Xian}
    \country{China} 
  }
\email{zxniu@xidian.edu.cn}

\author{Quan Wang}
\orcid{0000-0001-6913-8604}
\affiliation{%
  \institution{Xidian University}
      \city{Xian}
    \country{China} 
  }
\email{qwang@xidian.edu.cn}

\renewcommand{\shortauthors}{Haodi et al.}

\begin{abstract}
  Dynamic graph-level embedding aims to capture structural evolution in networks, which is essential for modeling real-world scenarios. However, existing methods face two critical yet under-explored issues: Structural Visit Bias, where random walk sampling disproportionately emphasizes high-degree nodes, leading to redundant and noisy structural representations; and Abrupt Evolution Blindness, the failure to effectively detect sudden structural changes due to rigid or overly simplistic temporal modeling strategies, resulting in inconsistent temporal embeddings. To overcome these challenges, we propose EvoFormer, an evolution-aware Transformer framework tailored for dynamic graph-level representation learning. To mitigate Structural Visit Bias, EvoFormer introduces a Structure-Aware Transformer Module that incorporates positional encoding based on node structural roles, allowing the model to globally differentiate and accurately represent node structures. To overcome Abrupt Evolution Blindness, EvoFormer employs an Evolution-Sensitive Temporal Module, which explicitly models temporal evolution through a sequential three-step strategy: (I) Random Walk Timestamp Classification, generating initial timestamp-aware graph-level embeddings; (II) Graph-Level Temporal Segmentation, partitioning the graph stream into segments reflecting structurally coherent periods; and (III) Segment-Aware Temporal Self-Attention combined with an Edge Evolution Prediction task, enabling the model to precisely capture segment boundaries and perceive structural evolution trends, effectively adapting to rapid temporal shifts. Extensive evaluations on five benchmark datasets confirm that EvoFormer achieves state-of-the-art performance in graph similarity ranking, temporal anomaly detection, and temporal segmentation tasks, validating its effectiveness in correcting structural and temporal biases. Code is available at \footnote{\url{https://github.com/zlx0823/EvoFormerCode}}.
\end{abstract}




\maketitle

\section{Introduction}

Dynamic graphs consist of nodes and edges whose topology evolves over discrete time intervals (e.g., edge additions/deletions). Formally, dynamic graphs are defined as a sequence of snapshots $G=\{G_t\}_{t=1}^T$, where each snapshot $G_t=(V_t, E_t)$ captures node interactions at time $t$. Dynamic graph-level representation seeks to derive compact, meaningful representations that capture both structural and temporal characteristics of the entire graph. These embeddings are fundamental to various downstream tasks, such as dynamic graph classification~\cite{zheng2025survey}, graph-level anomaly detection~\cite{goyal2018dyngem}, and graph similarity ranking ~\cite{bai2019unsupervised}.

Existing methods for graph representation predominantly focus on static graphs, aggregating node interactions without explicitly modeling temporal dynamics. Such static embedding approaches inherently disregard crucial temporal information, such as the evolution of nodes and edges, thus restricting their applicability to dynamic scenarios that demand precise temporal modeling~\cite{barros2021survey, goyal2018dyngem, beres2019node}. Recently, dynamic graph representation methods have emerged, predominantly emphasizing node-level embeddings, which typically encode individual nodes by explicitly incorporating node attributes, structural contexts, and temporal dependencies. Prominent node-level approaches integrate Graph Neural Networks (GNNs) with sequential models such as Recurrent Neural Networks (RNNs) or Transformer architectures~\cite{pareja2020evolvegcn, cong2023dyformer, wu2024feasibility}, employing mechanisms like recurrence or self-attention to effectively preserve node-level temporal dynamics. However, these node-level embeddings face inherent limitations when addressing tasks requiring holistic modeling of global graph structures and their temporal evolution. Tasks such as temporal graph-to-graph similarity assessment~\cite{bai2019unsupervised, beladev2020tdgraphembed}, retrieving~\cite{wang2016multimodal,wang2020joint}, graph clustering~\cite{zhong2022clustering,zhong2021clustering,zhong2020clustering}, temporal anomaly detection, and trend analysis~\cite{goyal2018dyngem, savage2014anomaly} explicitly demand global, graph-level embeddings. To derive graph-level representations from node embeddings, prior methods usually rely on indirect, simplistic aggregation strategies such as mean or sum pooling~\cite{duvenaud2015convolutional}, virtual node insertion~\cite{li2015gated}, or set-based pooling~\cite{gilmer2017neural}. However, these approaches inherently lack the capability to preserve global hierarchical structures and temporal semantics~\cite{ying2018hierarchical}, thereby significantly constraining their expressiveness in dynamic graph contexts. This limitation underscores the necessity of developing dedicated graph-level embedding approaches capable of accurately capturing both complex global structural patterns and evolving temporal dynamics.

Explicitly graph-level embedding methods for dynamic graphs remain relatively scarce. To our best knowledge, tdGraphEmbed~\cite{beladev2020tdgraphembed} was the first approach to explicitly generate temporal graph-level embeddings. Specifically, tdGraphEmbed extended random walk-based node embedding techniques to directly compute entire graph embeddings for each snapshot. However, tdGraphEmbed treated dynamic graphs merely as sequences of independent static snapshots, lacking comprehensive modeling of inter-snapshot temporal correlations. To better capture temporal dependencies, teleEmbed~\cite{huang2022learning} introduced temporal backtracking random walks, allowing the model to revisit previous temporal states. Nonetheless, this approach remains limited, as its random walk strategy can only sample local neighborhoods across time steps, thus restricting temporal modeling strictly to local structural contexts. Recently, GraphERT~\cite{beladev2023graphert} further advanced dynamic graph-level embedding by employing Transformer architectures to learn contextual node information directly from random walk sequences. Compared to tdGraphEmbed and teleEmbed, GraphERT utilizes a masking strategy that enables more precise modeling of local neighborhood contexts and predicts the temporal index of input random walk sequences to capture localized temporal evolution. Nevertheless, these advances remain confined to modeling temporal dynamics within local communities, and lack effective mechanisms for capturing global temporal evolution.

\begin{figure}[t]
  \centering

  \begin{subfigure}[t]{0.5\textwidth}
    \centering
    \includegraphics[width=\linewidth]{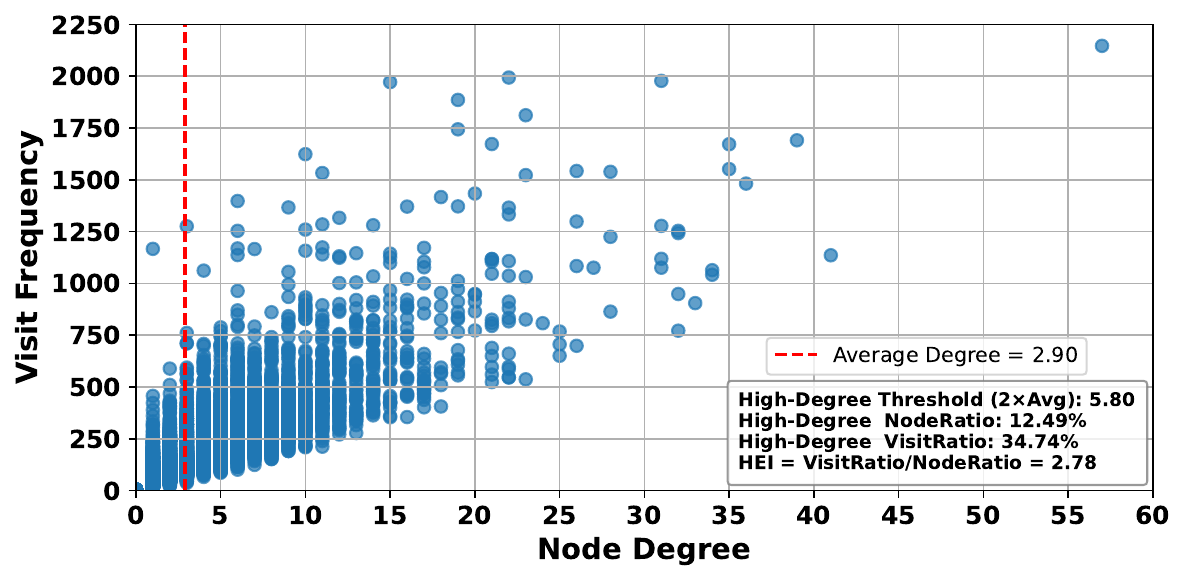}
    \label{fig:problem1}
  \end{subfigure}
  \vspace{1em}
  \vspace{-2.8em}
  \begin{subfigure}[t]{0.5\textwidth}
    \centering
    \includegraphics[width=\linewidth]{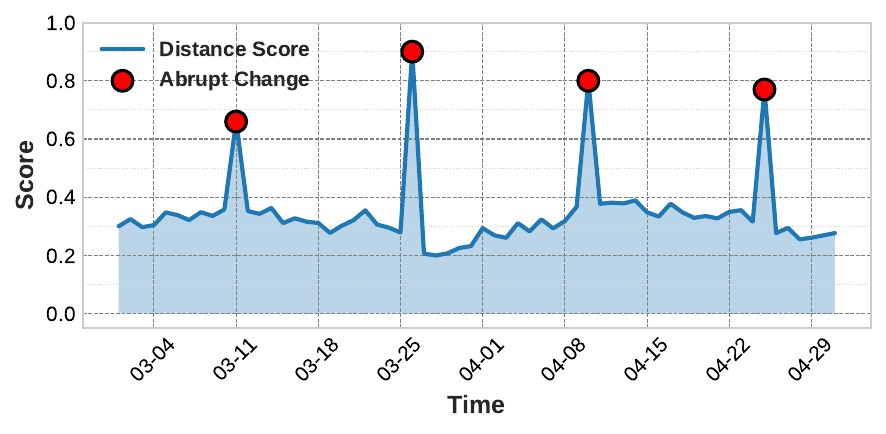}
    \label{fig:problem2}
  \end{subfigure}
 \vspace{-2.9em}
  \caption{Illustration of two key challenges in dynamic graph-level representation. The top figure shows \emph{Structural Visit Bias}. The bottom figure highlights \emph{Abrupt Evolution Blindness}.}
  \label{fig:combined}
\end{figure}

As a result, despite recent progress, existing dynamic graph-level embedding approaches still suffer from two critical yet under-explored limitations. Firstly, \emph{Structural Visit Bias (SVB)}, originating from random walk-based sampling, disproportionately emphasizes high-degree nodes, producing redundant and noisy structural representations. As shown in the top part of Fig.~\ref{fig:combined}, high-degree nodes constitute merely 12.49\% of all nodes but account for 34.66\% of total visits, yielding a high-degree exposure index (HEI) of 2.78. Such excessive visitation highlights significant structural bias, resulting in redundant sampling and semantic distortions. Secondly, existing approaches suffer from \emph{Abrupt Evolution Blindness (AEB)}, which describes the failure of dynamic graph models to accurately perceive and respond to sudden, significant structural changes during graph evolution. As illustrated in the bottom part of Fig.~\ref{fig:combined}, dynamic graphs typically exhibit non-uniform temporal patterns, characterized by long periods of gradual structural drift punctuated by rare but sharp transitions. These abrupt changes often correspond to real-world events, such as social unrest or system failures, that introduce semantically critical but temporally sparse structural mutations. However, current methods rely on fixed-size sliding windows or uniform temporal dependency assumptions. Such designs inherently smooth over or dilute these outlier events, resulting in semantic drift, where embeddings fail to reflect meaningful structural mutations at key timesteps. Consequently, the learned graph-level representations become temporally inconsistent and misaligned with the true evolution trajectory. This limits the model’s ability to generalize across distribution shifts and prevents accurate modeling of global temporal dynamics.

To overcome these challenges, we propose EvoFormer, an evolution-aware Transformer framework. EvoFormer comprises two modules designed to systematically mitigate structural and temporal biases. The first component, the Structure-Aware Transformer Module (SATM), explicitly injects node structural roles into the Transformer, globally differentiating structurally similar but semantically distinct nodes, thereby effectively alleviating Structural Visit Bias and minimizing redundant structural representations. Furthermore, EvoFormer integrates an Evolution-Sensitive Temporal Module (ESTM). It initially generates timestamp-aware embeddings; subsequently, a Graph-Level Temporal Segmentation step partitions the embedding sequence into structurally coherent intervals. Finally, EvoFormer employs Segment-Aware Temporal Self-Attention combined with an Edge Evolution Prediction task to capture comprehensive global temporal patterns and sensitively detect abrupt structural changes. As a result, EvoFormer produces temporally consistent embeddings that accurately reflect both gradual and abrupt changes.

The contributions of this work are threefold: (I) We formally identify and address two critical yet previously overlooked challenges in dynamic graph embedding: Structural Visit Bias and Abrupt Evolution Blindness; (II) We propose EvoFormer, which systematically integrates structure-aware positional encoding and temporal segmentation mechanisms to comprehensively mitigate these biases; (III) Comprehensive experiments on five benchmark datasets demonstrate EvoFormer's state-of-the-art performance in graph similarity ranking, temporal anomaly detection, and temporal segmentation tasks.

\section{Related Work}

\noindent{\bf Node Level Embedding.} These methods have gained considerable attention due to their effectiveness in modeling graph structures. For static graph embeddings, widely used techniques include matrix decomposition methods such as Singular Value Decomposition (SVD) \cite{belkin2001laplacian} and Locally Linear Embedding (LLE) \cite{roweis2000nonlinear}, as well as random walk-based approaches combined with the word2vec framework~\cite{mikolov2013efficient,grover2016node2vec}. Additionally, SDNE~\cite{wang2016structural} employs an autoencoder to simultaneously preserve first- and second-order proximities. Although these methods excel at capturing local structures, they typically neglect the global topology. GNNs \cite{kipf2016semi, velivckovic2018graph, hamilton2017inductive,kipf2016variational} overcome this limitation through message-passing mechanisms that aggregate information from local neighborhoods, effectively integrating both local and global contexts. However, deeper GNN architectures often encounter oversmoothing issues, causing node representations to become indistinct and hindering the discrimination capability between nodes~\cite{zhang2020graph}. Recent works thus integrate Transformer layers into GNNs by modifying core components such as feature aggregation strategies~\cite{cai2020graph, ying2021transformers}.

For dynamic graphs, node embedding methods usually extend static embedding approaches by independently embedding graph snapshots and then aligning these embeddings temporally~\cite{liang2021cross,singer2019node}. Other methods explicitly capture temporal dependencies. DynSDNE~\cite{goyal2018dyngem} initializes current embeddings based on historical representations, while DynAERNN~\cite{goyal2020dyngraph2vec} incorporates recurrent neural networks to model temporal dynamics. DynamicTriad~\cite{zhou2018dynamic} explicitly models network evolution through triadic closure processes. Similarly, tNodeEmbed~\cite{singer2019node} integrates historical embeddings for improved node-level predictions. EvolveGCN~\cite{pareja2020evolvegcn} dynamically adjusts GNN parameters via recurrent architectures like LSTM and GRU, effectively modeling graph evolution without explicit node embeddings. TREND~\cite{wen2022trend} employs a Hawkes-process-based GNN for modeling event-driven dynamics, though it faces scalability issues. Additionally, MTSN~\cite{liu2021motif} incorporates motif-based features within temporal shift mechanisms to capture intricate evolutionary patterns.

Despite their effectiveness, node-level embeddings typically rely on heuristic pooling strategies, including mean, sum, virtual-node, or set-based readouts~\cite{duvenaud2015convolutional, li2015gated, gilmer2017neural, ying2018hierarchical}, which discard hierarchical structure and essential temporal semantics. To overcome these limitations, recent research increasingly emphasizes graph-level embedding methods capable of naturally encoding complete structural and temporal dynamics.

\noindent{\bf Graph-Level Embedding.} These methods primarily aim to represent entire graphs as unified vectors. Static graph embedding methods include Graph2vec~\cite{narayanan2017graph2vec}, which treats graphs as documents and extracts embeddings from subgraph structures; UGraphEmb~\cite{bai2019unsupervised}, preserving similarities using graph edit distance; and Sub2vec~\cite{adhikari2018sub2vec}, learning subgraph embeddings through random walks. Other methods construct graph-level embedding based on degree histograms~\cite{cai2018simple}, spectral properties~\cite{galland2019invariant}, characteristic functions derived from node neighborhoods~\cite{rozemberczki2020characteristic}, and diffusion wavelets~\cite{wang2021graph}.

Dynamic graph embedding methods typically extend static frameworks temporally. tdGraphEmbed~\cite{beladev2020tdgraphembed} independently aggregates node embeddings per snapshot, ignoring temporal dependencies. teleEmbed~\cite{huang2022learning} introduces temporal-backtracking walks to capture inter-snapshot relations but relies on simple aggregation. Recently, GraphERT~\cite{beladev2023graphert} significantly advanced dynamic embeddings by integrating Transformer architectures to directly model structural and temporal contexts from time-stamped walks, establishing the current state-of-the-art.

However, existing methods remain limited by noisy structural representations from repeated hub-node visits and overly simplistic temporal modeling. To address these challenges, our method introduces structure-aware positional encodings and evolution-sensitive temporal module, effectively enhancing embedding robustness and temporal precision.

\begin{figure*}[htbp]
    \centering
    \includegraphics[width=\textwidth]{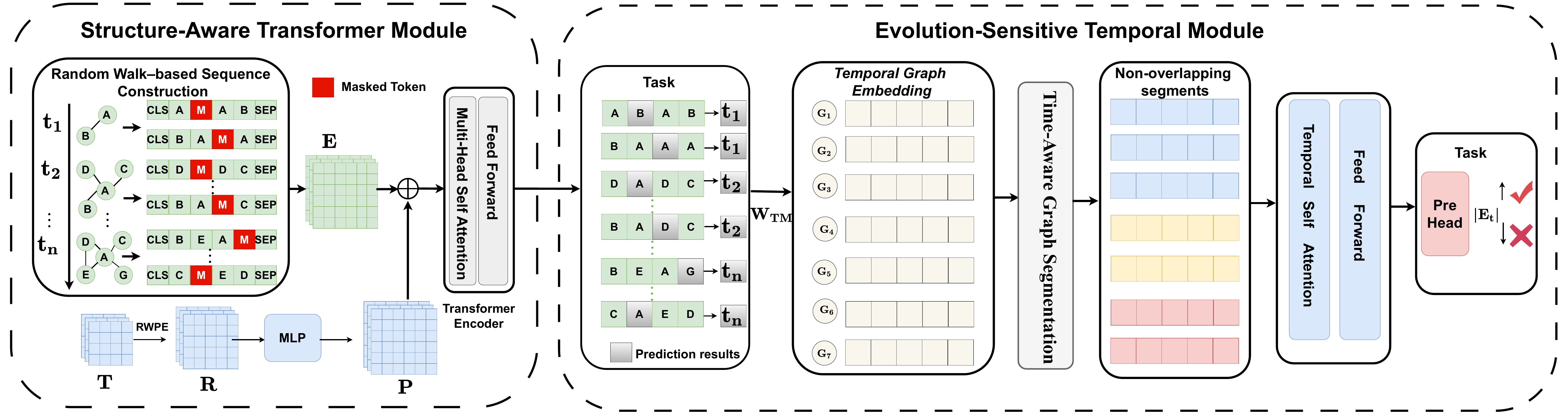} 
    \vspace{-2em}
    \caption{The overall architecture of EvoFormer. The framework consists of two modules: (1) the Structure-Aware Transformer Module encodes random-walk sequences using structure-aware positional encodings based on return probability vectors; (2) the Evolution-Sensitive Temporal Module performs timestamp prediction, dynamic graph segmentation, segment-aware temporal attention, and edge prediction to capture non-uniform graph evolution.}
    \label{fig: model1} 
\end{figure*}

\section{Problem statement}

\begin{problem}[Dynamic Graph-Level Representations (DGR)]  
Let $G = \{G_1, G_2, \ldots, G_T\}$ denote a dynamic graph, where each graph snapshot $G_t = (V_t, E_t)$ represents the node set $V_t$ and edge set $E_t$ at time $t$, $t\in[T]$. Let $V = \bigcup_{t=1}^{T} V_t$ denote the set of all nodes observed across all time steps, and $E = \bigcup_{t=1}^{T} E_t$ denote the set of all edges that appear in any snapshot.
The objective of DGR is to learn an embedding function $f: G_t \rightarrow \mathbf{g}_t \in \mathbb{R}^d$ for each snapshot $G_t$, such that the resulting sequence $\{\mathbf{g}_t\}_{t=1}^T
$ faithfully captures both the structural properties and temporal evolution of the underlying graph. Ideally, graph snapshots exhibiting similar topological and temporal patterns should be mapped to proximal locations in the embedding space.
\label{def:problem:DGR}
\end{problem}

\section{Proposed method}

This section introduces EvoFormer, a dynamic graph representation method that jointly addresses structural visit bias and abrupt evolution blindness. It integrates two core components: (I) a Structure-Aware Transformer Module, which encodes random-walk sequences enhanced by structure-aware positional encodings derived from return probability vectors; (II) an Evolution-Sensitive Temporal Module, which models non-uniform temporal evolution through timestamp classification, dynamic graph-level segmentation, segment-aware temporal self-attention, and edge prediction, thus accurately capturing abrupt structural transitions.

\subsection{Structure-Aware Transformer Module}

\noindent{\bf Random Walk–based Sequence Construction.} Given a temporal graph snapshot $G_t=(V_t, E_t)$, we construct a sequence-based representation of its structure by performing multiple biased random walks. For each node $v_i \in V_t$, $i \in [|V_t|]$, we initiate $W$ random walks of fixed length $L$~\cite{grover2016node2vec}, each defined as:
\begin{equation}
R_{v_i}^{(t, m)} = \{ R_{v_i}^1, R_{v_i}^2, \ldots, R_{v_i}^L \},~ \text{where } R_{v_i}^1 = v_i,~ t \in [T],~m \in [W].
\end{equation}

At each step $l \in [L - 1]$, the next node $R_{v_i}^{l+1}$ is sampled from the neighbors of $R_{v_i}^l$ according to the transition probabilities~\cite{grover2016node2vec}. Applying this procedure to every node across all $T$ graph snapshots results in a random walk corpus:
\begin{equation}
\mathcal{R} = \{ R_{v_i}^{(t, m)} \mid v_i \in V_t,\ t \in [T],\ m \in [W] \},
\end{equation}
This corpus $\mathcal{R}$ captures local structural patterns at each temporal step and serves as the input to the structure-aware Transformer module.

\noindent{\bf Structure-Aware Positional Encoding.} To alleviate \emph{Structural Visit Bias}, we incorporate structure-aware positional signals into node tokens using the \emph{Random Walk Positional Encoding} (RWPE) proposed by Dwivedi~\textit{et al.}~\cite{dwivedigraph}. Unlike naive positional embeddings ~\cite{vaswani2017attention}, RWPE captures the topological context of each node by quantifying its likelihood of returning to itself over multiple random walk steps, thereby encoding structural centrality and role beyond raw degree.

Formally, let $\mathbf{A}_t \in \{0, 1\}^{n \times n}$ denote the adjacency matrix of a temporal snapshot $G_t$, and let $\mathbf{D}_t = \mathrm{diag}(d_1, \ldots, d_n)$ be the diagonal degree matrix, and $n=|V_t|$. The one-step random walk transition matrix of $G_t$ is defined as:
\begin{equation}
\mathbf{T}_t = \mathbf{A}_t \mathbf{D}_t^{-1}
\label{eq:transition}
\end{equation}

For each node $v_i$, we compute its $k$-step \emph{return probability vector}:
\begin{equation}
    \mathbf{r}_i^t = \big[ (\mathbf{T}_t)_{ii}, (\mathbf{T}_t^2)_{ii}, \ldots, (\mathbf{T}_t^k)_{ii} \big],~ \mathbf{r}_i^t \in \mathbb{R}^k
    \label{eq:return_prob}
\end{equation}
where each entry $(\mathbf{T}_t^k)_{ii}$ represents the probability of returning to $v_i$ after $k$ random walk steps. This vector encodes the local diffusion behavior centered at $v_i$ and serves as a structural signature. Stacking these vectors for all nodes yields the positional encoding matrix:
\begin{equation}
\mathbf{R}^t
= \big[ \mathbf{r}_{1}^t, \mathbf{r}_{2}^t, \ldots, \mathbf{r}_{n}^t \big]^{\top}
 \in\mathbb{R}^{\,n\times k}
\label{eq:rwpe_matrix}
\end{equation}

By injecting $\mathbf{R}^t$ into the Transformer encoder, the model learns to differentiate structurally distinct nodes even when they occur in similar walk contexts. As a result, it mitigates redundancy and reduces the over-smoothing effects introduced by repeated visits to high-degree nodes.

\noindent{\bf Transformer-based Structural Context Modeling. }Given a random-walk sequence $R_{v_i}^{(t, m)} = \{ R_{v_i}^1, R_{v_i}^2, \ldots, R_{v_i}^L \}$, where $v_i \in V_t$, $t \in [T]$, and $m \in [W]$, we follow the BERT-style encoder architecture~\cite{devlin2019bert} to encode its contextual structure. We first prepend a special classification token \texttt{[CLS]} and append a \texttt{[SEP]} token to each sequence, yielding:
\begin{equation}
S_{v_i}^{(t,m)} = (\texttt{[CLS]}, R_{v_i}^{1}, R_{v_i}^{2}, \ldots,\texttt{[MASK]},\ldots, R_{v_i}^{L}, \texttt{[SEP]})
\end{equation}
with total sequence length $L' = L + 2$. During training, some positions are randomly replaced with \texttt{[MASK]} to enable masked language modeling.

To encode this sequence, we first embed each token using a learnable embedding matrix. Let $\mathcal{T}: \mathcal{V} \rightarrow {1, \ldots, |\mathcal{V}|}$ be a vocabulary lookup function mapping nodes and special tokens to unique token indices. The token embedding for the $l$-th position in the sequence is computed as:
\begin{equation}
\mathbf{e}_{l}=\mathbf{W}_{\text {tok }}[\mathcal{T}(S_{v_i}^{(t, m)}[l])],~ l \in [L^{\prime}~],~\mathbf{e}_{l}\in \mathbb{R}^d
\end{equation}
where $\mathbf{W}_{\text{tok}} \in \mathbb{R}^{|\mathcal{V}| \times d}$ is a learnable embedding matrix.

Stacking all embeddings gives the token embedding matrix:
\begin{equation}
\mathbf{E}_{v_i}^{(t, m)}=\left[\mathbf{e}_1, \mathbf{e}_2, \ldots, \mathbf{e}_{L'}\right]^{\top} \in \mathbb{R}^{L' \times d} .
\end{equation}

To incorporate structural role information and alleviate \emph{Structural Visit Bias}, we further add structure-aware positional encodings. For each token $\mathcal{T}(R_{v_i}^l)$ in the walk, we extract its return probability vector $\mathbf{r}_l^t \in \mathbb{R}^k$ from Eq.~\eqref{eq:return_prob}. Special tokens like \texttt{[CLS]}, \texttt{[MASK]}, and \texttt{[SEP]} are assigned zero vectors. Stacking these vectors yields:
\begin{equation}
\mathbf{R}_{v_i}^{(t, m)}=\left[\mathbf{r}_1^t, \mathbf{r}_2^t, \ldots, \mathbf{r}_{L^{\prime}}^t\right]^{\top} \in \mathbb{R}^{L^{\prime} \times k}
\end{equation}

We project each row into $d$-dimensional space using a two-layer MLP $f_\theta: \mathbb{R}^k \rightarrow \mathbb{R}^d$:
\begin{equation}
\mathbf{p}_{l}^t=f_\theta(\mathbf{r}_{l}^t)=\mathbf{W}_2 \operatorname{ReLU}(\mathbf{W}_1 \mathbf{r}_{l}^t+\mathbf{b}_1)+\mathbf{b}_2, \quad l\in[L^{\prime}]
\end{equation}

Let $\mathbf{P}_{v_i}^{(t,m)} \in \mathbb{R}^{L' \times d}$ be the resulting matrix. The final input to the Transformer encoder is the sum of the token and positional embeddings:
\begin{equation}
\mathbf{X}_{v_i}^{(t, m)}=\mathbf{E}_{v_i}^{(t, m)}+\mathbf{P}_{v_i}^{(t, m)} \in \mathbb{R}^{L^{\prime} \times d}
\end{equation}

We feed $\mathbf{X}_{v_i}^{(t,m)}$ into a stack of $Z$ Transformer layers. Let $\mathbf{H}^{(0)} = \mathbf{X}_{v_i}^{(t,m)}$ and $\mathbf{H}^{(z)}$ denote the output of the $z$-th layer, $z\in[Z]$:
\begin{equation}
\mathbf{H}^{(z)}=\operatorname{TransformerLayer}(\mathbf{H}^{(z-1)})
\label{eq:Tout}
\end{equation}

We denote the final output by $\mathbf{H}_{v_i}^{(t,m)} = \mathbf{H}^{(Z)} \in \mathbb{R}^{L' \times d}$. The output corresponding to the masked positions is used for reconstruction. Let $\mathbf{h}_l$ be the representation at a masked position $l$, its token prediction is given by:
\begin{equation}
P(S_{v_i}^{(t, m)}[l]=v)=\operatorname{softmax}(\mathbf{W}_{\text {p}} \mathbf{h}_{l}+\mathbf{b}_{\text {p}}),
\end{equation}
where $\mathbf{W}_{\text{p}} \in \mathbb{R}^{|\mathcal{V}| \times d}$ and $\mathbf{b}_{\text{p}} \in \mathbb{R}^{|\mathcal{V}|}$ are trainable parameters.

\subsection{Evolution-Sensitive Temporal Module}

\noindent{\bf Random Walk Timestamp Classification.} Following GraphERT~\cite{beladev2023graphert}, we incorporate a timestamp prediction task to model local temporal dynamics. For each  $S_{v_i}^{(t, m)}$, we extract the contextualized \texttt{[CLS]} token embedding from the Transformer output:
\begin{equation}
\mathbf{h}_{\mathrm{CLS}, i}^{(t, m)}=\mathbf{H}_{v_i}^{(t, m)}[1] \in \mathbb{R}^d,
\end{equation}
where $\mathbf{H}_{v_i}^{(t, m)}$ is the final-layer hidden state matrix from Eq.~(\ref{eq:Tout}). We feed $\mathbf{h}_{\text {CLS }, i}^{(t, m)}$ into a linear classifier to predict the timestamp:
\begin{equation}
\hat{\mathbf{y}}_i^{(t, m)}=\operatorname{softmax}\left(\mathbf{W}_{\mathrm{TM}} \mathbf{h}_{\mathrm{CLS}, i}^{(t, m)}+\mathbf{b}_{\mathrm{TM}}\right) \in \mathbb{R}^T,
\end{equation}
where $\mathbf{W}_{\mathrm{TM}} \in \mathbb{R}^{T \times d}$ and $\mathbf{b}_{\mathrm{TM}} \in \mathbb{R}^T$ are learnable parameters.

This task forces the model to infer temporal context from structural patterns alone, enhancing its sensitivity to local temporal variations across snapshots.

\begin{algorithm}[t]\footnotesize
\caption{\textsc{Top-Down Segmentation}$(\mathbf{W}_{\mathrm{TM}}, p)$}
\label{alg:topdown}
\begin{algorithmic}[1]
\Require $\mathbf{W}_{\mathrm{TM}} \in \mathbb{R}^{T \times d}$: graph-level embeddings for $T$ time steps; $p$: number of desired segments
\Ensure Segmentation vector $\mathbf{v} \in \{1,\dots,p\}^T$, where $\mathbf{v}[t]$ is the segment index of time step $t$

\State Initialize $\mathcal{S} \gets \{[1, T]\}$ \Comment{Start with full time span}
\While{$|\mathcal{S}| < p$}
    \State Select $[s, e] \in \mathcal{S}$ with largest length
    \State Initialize $j^* \gets -1$, $\text{best\_score} \gets -\infty$
    \For{$j = s+1$ to $e-1$}
        \State Compute summary vectors:
        \[
        \mathbf{s}_1 = \text{mean}(\mathbf{W}_{\mathrm{TM}}[s:j,:]),\quad \mathbf{s}_2 = \text{mean}(\mathbf{W}_{\mathrm{TM}}[j+1:e,:])
        \]
        \State Compute cosine similarity score:
        \[
        \text{score}_1 = \sum_{t=s}^{j} \cos(\mathbf{w}_t, \mathbf{s}_1),\quad \text{score}_2 = \sum_{t=j+1}^{e} \cos(\mathbf{w}_t, \mathbf{s}_2)
        \]
        \If{$\text{score}_1 + \text{score}_2 > \text{best\_score}$}
            \State $j^* \gets j$, $\text{best\_score} \gets \text{score}_1 + \text{score}_2$
        \EndIf
    \EndFor
    \State Replace $[s, e]$ in $\mathcal{S}$ with $[s, j^*]$ and $[j^*+1, e]$
\EndWhile
\State Initialize $\mathbf{v} \gets \mathbf{0} \in \mathbb{R}^{T}$
\For{$i = 1$ to $p$}
    \State Let $[s, e]$ be the $i$-th segment in $\mathcal{S}$
    \For{$t = s$ to $e$}
        \State $\mathbf{v}[t] \gets i$
    \EndFor
\EndFor
\State \Return $\mathbf{v}$
\end{algorithmic}
\end{algorithm}

\noindent{\bf  Time-Aware Graph Segmentation. } To enhance global temporal modeling and better capture non-uniform structural evolution, we apply a top-down segmentation strategy~\cite{terzi2006efficient} over graph-level embeddings. Standard self-attention often assigns disproportionate weights to distant time steps, which dilutes the impact of local transitions and hinder responsiveness to recent structural changes~\cite{xing2024less}. To address this, we partition the timeline into semantically coherent segments and restrict attention computation within each segment.

Given graph embeddings $\mathbf{W}_{\mathrm{TM}} \in \mathbb{R}^{T \times d}$ over $T$ time steps, we recursively divide the time axis into $p$ disjoint intervals that maximize intra-segment coherence. Each split is chosen to maximize the sum of cosine similarities between embeddings and their respective segment-level mean vectors (see Algorithm~\ref{alg:topdown}).\footnote{Each segmentation step considers $\mathcal{O}(T)$ candidate splits and computes $\mathcal{O}(Td)$ cosine similarities, yielding an overall complexity of $\mathcal{O}(Tpd)$.}

The resulting segmentation vector $\mathbf{v}$ is later used in the Segment-Aware Temporal Self-Attention module to constrain temporal attention within each segment. This reduces noise from unrelated intervals and improves sensitivity to phase-specific transitions, especially abrupt changes.

\begin{figure}[htbp]
    \centering
    \includegraphics[scale = 0.35]{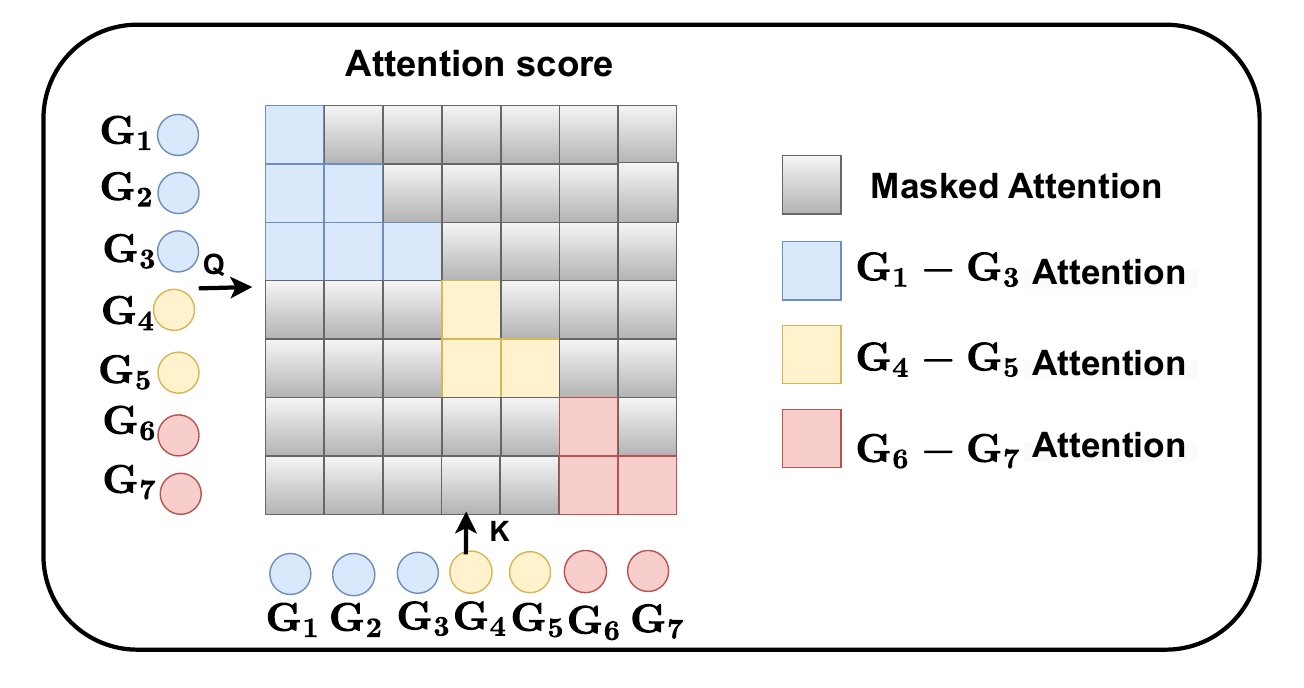}
    \vspace{-1.7em}
    \caption{Segment-Aware Temporal Self-Attention: each graph attends only to others within the same temporal segment, while attention to other segments is masked.}
    \label{fig:mode2}
\end{figure}

\noindent{\bf Segment-Aware Temporal Self-Attention. }To model global temporal dependencies while respecting temporal locality, we apply a modified temporal self-attention mechanism (inspired by ~\cite{sankar2018dynamic}) over the graph-level embedding matrix $\mathbf{W}_{\mathrm{TM}} \in \mathbb{R}^{T \times d}$ (See Fig.~\ref{fig:mode2}). We also utilize the segmentation vector $\mathbf{v} \in \{1, \ldots, p\}^T$ generated by Algorithm~\ref{alg:topdown}, where $\mathbf{v}[i]$ denotes the segment index of time step $i$.

The attention output $Z_g \in \mathbb{R}^{T \times d}$ is computed as:
\begin{equation}
Z_g = \beta_g \cdot (\mathbf{W}_{\mathrm{TM}} W_V),
\end{equation}
where $W_V \in \mathbb{R}^{d \times d}$ is the learnable value projection matrix, and $\beta_g \in \mathbb{R}^{T \times T}$ denotes the segment-aware attention weight matrix. Specifically, for each time step $i$, the attention weight $\beta_g^{ij}$ assigned to time step $j$ is computed as:
\begin{equation}
\beta_g^{ij} = \frac{\exp(e_g^{ij})}{\sum_{ t=1}^{T} \exp(e_g^{it})},
\end{equation}
where the attention score $e_g^{ij}$ is defined by:
\begin{equation}
e_g^{ij} = \frac{(\mathbf{W}_{\mathrm{TM}} W_Q)(\mathbf{W}_{\mathrm{TM}} W_K)^\top_{ij}}{\sqrt{d}} + M_{ij}.
\end{equation}

Here, $W_Q, W_K \in \mathbb{R}^{d \times d}$ are the learnable query and key projection matrices, and $M \in \mathbb{R}^{T \times T}$ is a segment-aware causal mask given by:
\begin{equation}
M_{ij} =
\begin{cases}
0, & \text{if } i \le j \text{ and } \mathbf{v}[i] = \mathbf{v}[j], \\
-\infty, & \text{otherwise}.
\end{cases}
\end{equation}

This masking enforces two constraints: (I) each time step $i$ can only attend to itself and earlier time steps $j \le i$; and (II) attention is restricted within the same segment, i.e., $\mathbf{v}[i] = \mathbf{v}[j]$. Together, these constraints ensure that the attention mechanism captures meaningful intra-phase temporal dependencies.

\noindent{\bf Edge Evolution Prediction.} To further enhance the model’s sensitivity to abrupt changes, we introduce an auxiliary prediction task that estimates edge count fluctuations across time steps. This component complements the Segment-Aware Temporal Self-Attention by explicitly supervising the model to recognize discrete transitions in graph dynamics.

Given the output of the temporal attention layer, $\mathbf{W}_{\mathrm{TM}} \in \mathbb{R}^{T \times d}$, where each row $\mathbf{w}_{\mathrm{TM}}^t$ represents the embedding of the graph at time step $t$, we form concatenated representations for each consecutive pair of timesteps:
\begin{equation}
\mathbf{z}^{t} = [\mathbf{w}^{t-1}_{\mathrm{TM}}|| \mathbf{w}^{t}_{\mathrm{TM}}], \quad t = [2, \dots, T].
\end{equation}

Each $\mathbf{z}^t \in \mathbb{R}^{2 d}$ is then fed into a binary classifier $f(\cdot)$ to predict whether the number of edges increases at time $t$ :
\begin{equation}
\hat{y}^{t} = \sigma(f(\mathbf{z}^{t})),
\end{equation}
where $\sigma(\cdot)$ denotes the sigmoid function, and $\hat{y}^t \in[0,1]$ denotes the predicted probability of an edge count increase between time steps $t-1$ and $t$.

\subsection{Joint Optimization}
Our model is optimized using a multi-task objective comprising three components: masked language modeling, random walk timestamp classification, and edge evolution prediction.

To reconstruct masked nodes within random walks, we apply a standard cross-entropy loss:
\begin{equation}
\mathcal{L}_{1} = -\frac{1}{|\mathcal{R}|} \sum_{i=1}^{|\mathcal{R}|} \sum_{(l, v)} \log\left((W_{\text{p}} \mathbf{h}_l)_v\right),
\end{equation}
where $\mathcal{R}$ is the training set, $\mathbf{h}_l$ is the hidden representation at a masked position $l$, and $(W_{\text{p}} \mathbf{h}_l)_v$ is the predicted probability of the true token $v$.

To capture local temporal context, we classify each random walk to its originating timestamp using the [CLS] token embedding:
\begin{equation}
\mathcal{L}_{2} = -\frac{1}{|\mathcal{R}|} \sum_{i=1}^{|\mathcal{R}|} \log\left((W_{\text{TM}} \cdot \mathbf{h}^{(t,m)}_{\text{CLS},i})_t\right),
\end{equation}
where $\mathbf{h}^{(t,m)}_{\text{CLS},i}$ is the contextualized [CLS] vector for the $i$-th sample, and $t$ is the ground-truth timestamp.

To encourage temporal sensitivity to structural changes, we adopt a binary classification loss to predict whether the edge count increases between timesteps:
\begin{equation}
\mathcal{L}_{3} =
-\sum_{t=2}^{T} \left[
y^{\,t}\log \hat{y}^{\,t} + (1 - y^{\,t}) \log (1 - \hat{y}^{\,t})
\right],
\end{equation}
where $y^{\,t} \in \{0, 1\}$ indicates edge growth between $G_{t-1}$ and $G_t$, and $\hat{y}^{\,t} \in [0, 1]$ is the predicted probability.

The total loss is a weighted combination of the three components:
\begin{equation}
\mathcal{L}_{\text{total}} = \lambda_1 \mathcal{L}_{1} + \lambda_2 \mathcal{L}_{2} + \lambda_3 \mathcal{L}_{3},
\end{equation}
where $\lambda_1$, $\lambda_2$, and $\lambda_3$ control the relative importance of each task.

\begin{table}[htbp]
  \setlength{\tabcolsep}{2pt}
  \renewcommand{\arraystretch}{0.85}
  \caption{Datasets Statistics.}
  \vspace{-1.5em}
  \label{tab:datasets}
  \begin{tabular}{lccccc}
    \toprule
\textbf{Datasets}&\textbf{Nodes}&\textbf{Edges}&\textbf{Timestamps}& \textbf{Time Resolution}\\
    \midrule
    FB & 46,873 & 857,815 & 30 & Monthly \\
    ER & 87,062 & 1,146,800 & 42 & Monthly \\
    R-GOT & 156,732 & 834,753 & 62 & Daily \\
    R-F1 & 38,702 & 254,731 & 61 & Daily \\
    DBLP$_{1}$ & 554 & 3,476 & 25 & Yearly\\
    DBLP$_{2}$ & 483 & 2,177 & 30 & Yearly\\
    DBLP$_{3}$ & 495 & 2,251 & 30 & Yearly\\
    DBLP$_{4}$ & 644 & 2,657 & 26 & Yearly\\
    DBLP$_{5}$ & 488 & 2,749 & 38 & Yearly\\
  \bottomrule
\end{tabular}
\end{table}

\section{Experimental Evaluation}

\subsection{Experimental Setup}

\noindent{\bf Datasets.} We evaluate EvoFormer on five publicly available large-scale dynamic graph datasets, as summarized in Table~\ref{tab:datasets}. Specifically, Facebook (FB), Enron (ER), Reddit 'Game of Thrones' (R-GOT), and Reddit 'Formula1' (R-F1) \cite{beladev2020tdgraphembed,huang2022learning,beladev2023graphert} are used to assess performance on graph similarity ranking and temporal anomaly detection tasks, while DBLP$_{1}$ through DBLP$_{5}$ \cite{zhong2024ego,zhong2022jaccard} are used to evaluate temporal segmentation.

\noindent{\bf Configuration Settings.} EvoFormer is trained for 15 epochs using the Adam optimizer with a fixed learning rate of $1\times10^{-4}$ and a batch size of 32. In the random walk-based sequence construction module, each node generates $W = 5$ walks of length $L = 32$. Return probability vectors are computed with $k = 16$ steps. The Transformer encoder consists of $Z = 8$ layers, each with hidden dimension $d = 256$ and $8$ attention heads. For temporal segmentation, the number of segments is set to $p = 8$. The total loss is a weighted combination of three objectives, with weights $\lambda_1 = 5$, $\lambda_2 = 10$, and $\lambda_3 = 5$. All experiments are conducted using Python 3.8 and PyTorch 2.2.1 on a single NVIDIA GeForce RTX 4090 GPU.

\noindent{\bf Baselines.} We compare EvoFormer with a wide range of node-level and graph-level embedding baselines\footnote{To ensure fair comparison, we run all baseline methods using their official open-source implementations and default hyperparameter settings. We report the best results obtained under these configurations.}. For node-level methods, we obtain graph-level representations by averaging node embeddings across the graph. To compensate for the lack of temporal modeling in static methods, we apply temporal alignment following~\cite{hamilton2017representation}.

\emph{Node-level embedding methods.} We consider both static and dynamic node embedding baselines.  
\textbf{Node2Vec}~\cite{grover2016node2vec}, \textbf{SDNE}~\cite{wang2016structural}, and \textbf{GAE}~\cite{kipf2016variational} are classical static methods that learn node representations via biased random walks, deep autoencoders, and encoder-decoder architectures, respectively.  
\textbf{DynamicTriad}~\cite{zhou2018dynamic}, \textbf{Dynamic SDNE}~\cite{goyal2018dyngem}, and \textbf{DynAERNN}~\cite{goyal2020dyngraph2vec} incorporate temporal dynamics through triadic closure modeling, proximity preservation, and RNN-based autoencoding.  
\textbf{EvolveGCN}~\cite{pareja2020evolvegcn} adapts GCN parameters over time via recurrent units.  
\textbf{MTSN}~\cite{liu2021motif} integrates motif-aware features and temporal shifts, while \textbf{TREND}~\cite{wen2022trend} models fine-grained temporal events using Hawkes processes.

\emph{ Graph-level embedding methods.} These methods directly learn representations for entire graphs.  
\textbf{Graph2Vec}~\cite{narayanan2017graph2vec}, \textbf{UGraphEmb}~\cite{bai2019unsupervised}, and \textbf{Sub2Vec}~\cite{adhikari2018sub2vec} capture global or subgraph-level patterns using WL kernels, proximity preservation, and neighborhood contexts.  
\textbf{LDP}~\cite{cai2018simple}, \textbf{IGE}~\cite{galland2019invariant}, and \textbf{FEATHER-G}~\cite{rozemberczki2020characteristic} generate graph embedding based on degree histograms, spectral features, and characteristic functions over random walks.  
\textbf{D-Wavelets}~\cite{wang2021graph} encodes structural information using diffusion wavelets.  
Temporal graph baselines include \textbf{tdGraphEmbed}~\cite{beladev2020tdgraphembed} and \textbf{teleEmbed}~\cite{huang2022learning}, which perform time-aware random walks, and \textbf{GraphERT}~\cite{beladev2023graphert}, which applies masked Transformer modeling with temporal classification.

\subsection{Experimental Tasks}

We evaluate EvoFormer on three graph-level tasks: graph similarity ranking, temporal anomaly detection, and temporal segmentation.

\noindent{\bf Task 1: Graph Similarity Ranking.} This task evaluates whether graph-level embeddings preserve semantic similarity. Given a query graph, the model ranks a set of candidate graphs based on embedding similarity. We adopt Maximum Common Subgraph (MCS) as the ground truth similarity measure. A model performs well if its embedding-based ranking aligns with MCS. \textcolor{black}{We evaluate performance using three standard metrics~\cite{beladev2023graphert}: Precision@K, Mean Reciprocal Rank (MRR), and Mean Average Precision@K (MAP@K). Precision@K measures the proportion of relevant results within the top‑K retrieved items. MRR evaluates the inverse rank of the first relevant result, averaged over all queries. MAP@K captures the average precision of all relevant results within the top‑K positions, averaged across queries.}

\noindent{\bf Task 2: Temporal Anomaly Detection.} This task aims to identify time steps where the graph exhibits significant structural deviations from typical behavior, such as abrupt changes, community shifts, or the emergence of new hubs. 
\textcolor{black}{For each time step, we compute an anomaly score based on the average classification confidence of path‑level temporal predictions~\cite{beladev2023graphert}.}

We evaluate performance on two benchmark datasets: \textsc{R-GOT} and \textsc{R-F1}. We report Mean Reciprocal Rank (MRR) to assess the model's ability to rank true anomalies among all time steps. To further evaluate the correlation between the detected anomalies and real-world trends, we report the Spearman correlation coefficient between the anomaly score sequence and the corresponding Google Trends time series. The Spearman correlation measures the monotonic relationship between two variables by comparing the rankings of their values, thus quantifying how well the model's anomaly detections align with fluctuations in public interest.

\noindent{\bf Task 3: Temporal segmentation. }This task aims to segment a dynamic graph sequence into distinct phases, each corresponding to a structurally coherent time interval. We conduct experiments on the DBLP dataset, which contains five temporal graph sequences annotated with ground-truth segment labels \cite{zhong2022jaccard,zhong2024ego}. For each sequence, we first compute graph-level embeddings, then apply binary segmentation algorithm based on dynamic programming over the embedding trajectories. This procedure evaluates whether the learned representations capture meaningful temporal shifts in graph structure. Performance is measured using three standard clustering metrics: accuracy (ACC), normalized mutual information (NMI), and macro-averaged F1 score (F1-macro), which collectively assess the alignment between predicted segments and ground-truth segment labels.

\begin{table}[htbp]
  \centering
  \small
  \caption{Performance on Graph Similarity Ranking. The best results are in bold. The second-best results are underlined.}
  \label{tab:ex1}
  \vspace{-1.7em}
  \setlength{\tabcolsep}{1pt}
  \renewcommand{\arraystretch}{0.9}
\begin{tabular}{c|cccc|cccc}
\toprule
\multirow{2}{*}{\textbf{Model}} & \multicolumn{4}{c|}{ \textbf{FB} } & \multicolumn{4}{c}{ \textbf{ER} } \\ \cmidrule{2-9}
& P@5 & P@10 & MRR & MAP@10  & P@5 & P@10 & MRR & MAP@10 \\
\midrule 
 Node2vec & 0.600 & 0.670 & 0.416 & 0.779 & 0.285 & 0.366 & 0.272 & 0.478  \\
 SDNE & 0.220 & 0.393 & 0.202 & 0.402 & 0.043 & 0.093 & 0.048 & 0.091  \\
 GAE & 0.480 & 0.603 & 0.373 & 0.727 & 0.237 & 0.288& 0.166 & 0.395 \\
 DynamicTriad & 0.677 & 0.743 &  \underline{0.483} & 0.840 & 0.234 & 0.323 & 0.250  & 0.408  \\
 Dynamic SDNE & 0.386 & 0.550 & 0.207 & 0.534  & 0 & 0.005 & 0.022 & 0  \\
 DynAERNN & 0.220 & 0.370 & 0.177 & 0.372  &  0.024 & 0.051 & 0.030 & 0.054\\
 EvolveGCN & 0.560 & 0.703 & \textbf{0.592}& 0.349  &0.376 & 0.431 & 0.313 & 0.506\\
 MTSN & 0.158 & 0.297 & 0.123& 0.327  &0.054 & 0.116 & 0.042 & 0.157\\
 TREND & 0.64 & 0.723 & 0.383& 0.833  &0.197 & 0.274 & 0.129 & 0.274\\

 \midrule 
 Graph2vec & 0.220 & 0.433 & 0.162 & 0.506 &0.032& 0.041 & 0.038 & 0.069 \\
 UGraphEmb & 0.567 & 0.750 & 0.268 & 0.815 &0.107& 0.162 & 0.122 & 0.204 \\
 Sub2Vec & 0.293 & 0.557 & 0.194 & 0.476 &0.042& 0.055 & 0.089 & 0.102 \\
 LDP & 0.387 & 0.517 & 0.189 & 0.606  & 0.290 & 0.474& 0.319 &0.550 \\
 IGE & 0.193 & 0.360 &0.182 & 0.411  & 0.138 & 0.274& 0.154 &0.318 \\

 FEATHER-G  & 0.253 & 0.407 & 0.196 & 0.429 & 0.319 &0.452 &0.217  &0.499 \\
 D-Wavelets & 0.333 & 0.510 & 0.228& 0.602 & 0.290 &0.462 &0.247 & 0.493  \\
 tdGraphEmbed & 0.660 & 0.760 & 0.392 & 0.831 & 0.272 & 0.369 & 0.306 & 0.445\\
 teleEmbed & 0.467 & 0.610 & 0.202 & 0.683 & 0.248&0.376 &0.151 & 0.433\\
 GraphERT & \underline{0.7} & \underline{0.773} & 0.508 & \underline{0.841} & \underline{0.523} & \underline{0.5} & \underline{0.645} & \underline{0.638}\\
\midrule EvoFormer  & \textbf{0.707} & \textbf{0.78} & 0.453 & \textbf{0.853}  & \textbf{0.585} & \textbf{0.547 }& \textbf{0.703} & \textbf{0.673}\\
\bottomrule
\toprule

\multirow{2}{*}{\textbf{Model}} & \multicolumn{4}{c|}{ \textbf{R-GOT} } & \multicolumn{4}{c}{ \textbf{R-F1} } \\ \cmidrule{2-9}
& P@5 & P@10 & MRR & MAP@10  & P@5 & P@10 & MRR & MAP@10 \\
\midrule 
Node2vec & 0.310 & 0.338 & 0.358 & 0.542 &0.167 & 0.206 & 0.204 & 0.376 \\
SDNE & 0.252 & 0.346 & 0.305 & 0.393 &0.151 & 0.231 & 0.201 & 0.273 \\
GAE & 0.224 & 0.267 & 0.249 & 0.409  & 0.141 & 0.204 & 0.195 & 0.333  \\
DynamicTriad & 0.282 & 0.301 & 0.402 & 0.487 &0.190 & 0.231 & 0.208 & 0.395 \\
Dynamic SDNE & 0.300 & 0.316 & 0.330 & 0.494 &0.141 & 0.201 & 0.101 & 0.289 \\
DynAERNN & 0.087& 0.166 & 0.088& 0.196  &0.095 & 0.203 & 0.081 & 0.255\\
EvolveGCN & 0.138 & 0.211 & 0.107& 0.267  &0.203 & 0.257 & 0.166 & 0.366\\
MTSN & 0.097 & 0.162 & 0.074& 0.221  &0.093 & 0.203 & 0.131 & 0.224\\
TREND & 0.180 & 0.295 & 0.192& 0.355  &0.121 & 0.226 & 0.109 & 0.282\\

\midrule
Graph2vec & 0.190 & 0.274 & 0.185  & 0.329  & 0.082 & 0.168 & 0.059 & 0.215  \\
UGraphEmb & 0.245 & 0.288 & 0.225 & 0.413 &0.170& 0.256 & 0.122 & 0.308 \\
Sub2Vec & 0.087 & 0.193 & 0.124 & 0.168 &0.032& 0.114 & 0.042 & 0.149 \\
LDP  & 0.210 & 0.285 & 0.178 & 0.341  & 0.157 & 0.233& 0.127 &0.304 \\
IGE  &0.083  & 0.176 & 0.101 & 0.202 & 0.172& \textbf{0.372}& 0.208 &0.360\\
FEATHER-G  & 0.219 & 0.255 & 0.158 & 0.348 & 0.147 &0.229 & 0.073  &0.282 \\
D-Wavelets & 0.226 & 0.279 & 0.212& 0.378& 0.164 & 0.242 & 0.100 & 0.311\\
tdGraphEmbed & 0.351 & 0.358 & 0.450 & 0.561  & \underline{0.229} & 0.273 & \textbf{0.257} & \underline{0.429}\\

teleEmbed & 0.077 & 0.143 & 0.069 & 0.174 &0.134 & 0.210&  0.103&  0.296\\
GraphERT  & \underline{0.416} & \underline{0.417} & \underline{0.479 }& \underline{0.582} &0.219 & 0.272 & \underline{0.239} & \textbf{0.46 }\\
\midrule EvoFormer  & \textbf{0.445} & \textbf{0.434} & \textbf{0.497 }& \textbf{0.622}  &\textbf{0.229} & \underline{0.293} & 0.211 & 0.402 \\
\bottomrule
\end{tabular}
\end{table}

\subsection{Experimental Results}

\begin{figure}[t]
    \centering
    \begin{subfigure}[b]{0.23\textwidth}
        \includegraphics[width=\textwidth]{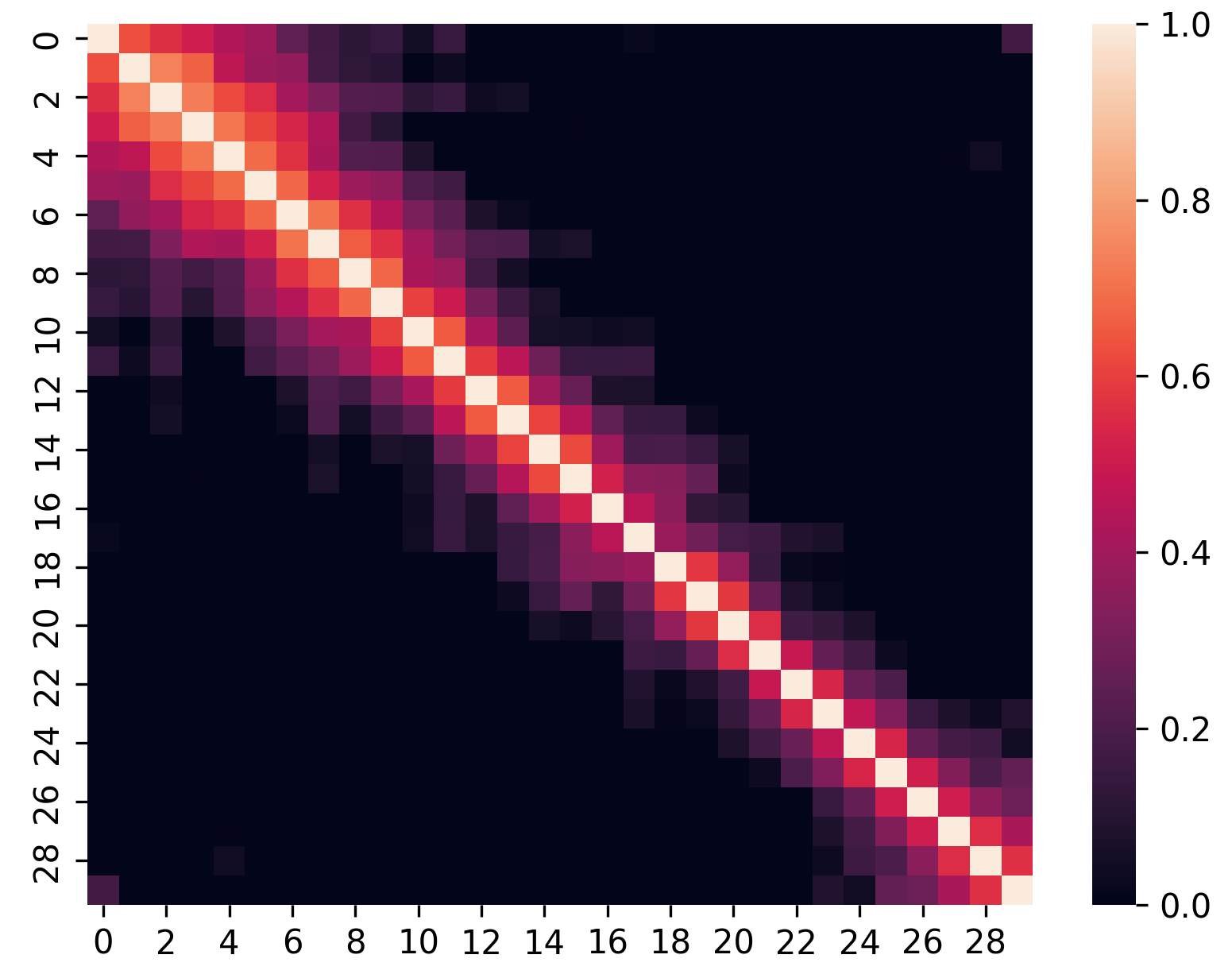}\vspace{-2.5mm}
        \caption{}
    \end{subfigure}
    \hfill
    \begin{subfigure}[b]{0.23\textwidth}
        \includegraphics[width=\textwidth]{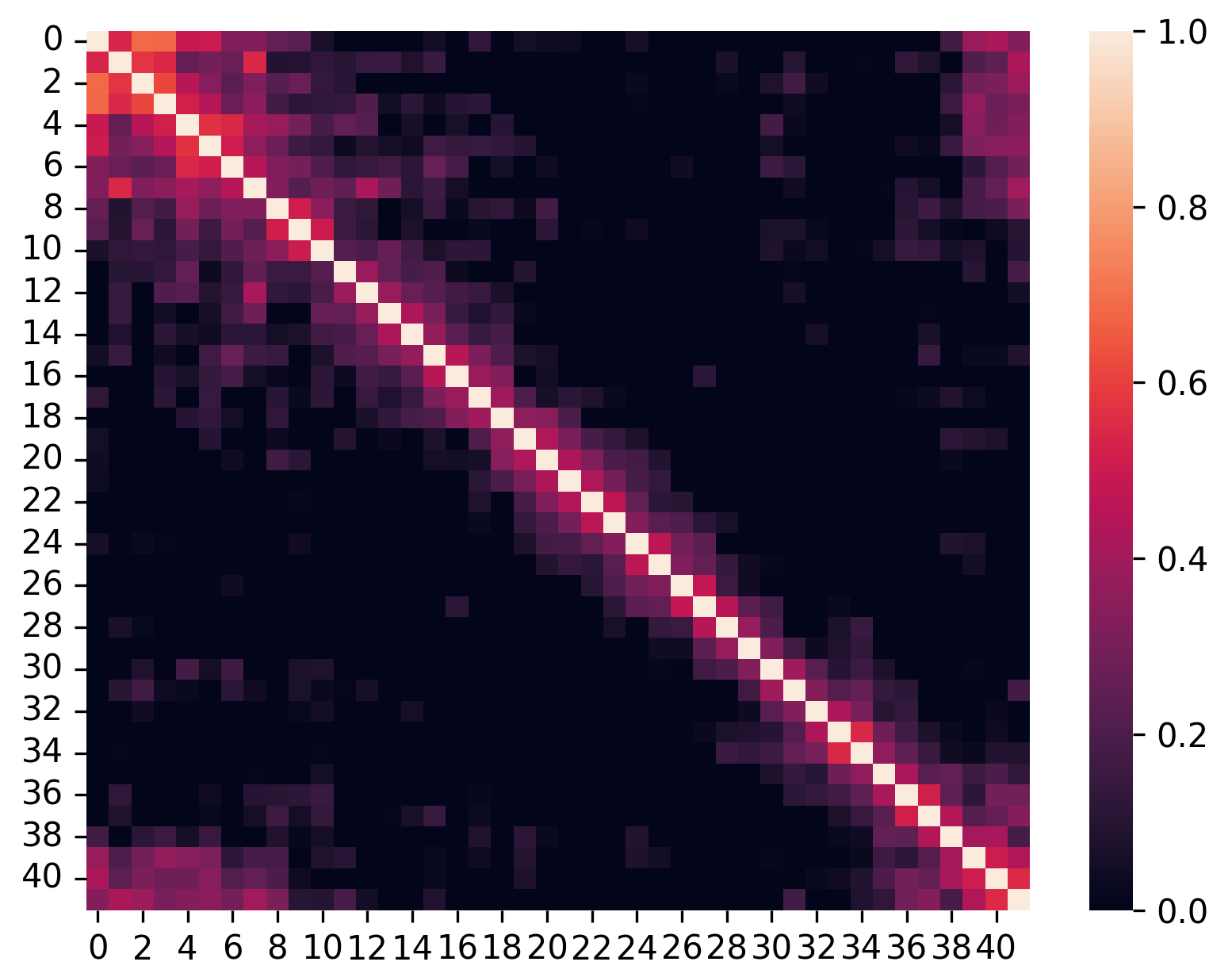}\vspace{-2.5mm}
        \caption{}
    \end{subfigure}
    \vspace{-1.5em}
    \caption{Graph-to-graph similarity heatmaps based on cosine similarity between graph embeddings over time. (a) Facebook: smooth evolution. (b) Enron: abrupt structural shifts.}
    \label{fig:heatmaps}
\end{figure}

\begin{table}[b]
\caption{MRR and Spearman results for temporal anomaly detection on R-GOT and R-F1 datasets.}
 \vspace{-1.4em} 
\label{tab:ex2}
\centering
\setlength{\tabcolsep}{2pt} 
\renewcommand{\arraystretch}{0.9} 
\begin{tabular}{l|c|c|c|c}
\toprule
\textbf{Model} & \multicolumn{2}{c|}{\textbf{R-GOT}} & \multicolumn{2}{c}{\textbf{R-F1}} \\
\cmidrule(lr){2-3} \cmidrule(lr){4-5}
 & \textbf{MRR} & \textbf{Spearman} & \textbf{MRR} & \textbf{Spearman} \\
\midrule
Node2vec      & \textbf{0.370} & 0.559 & 0.129 & 0.449 \\
SDNE          & 0.034 & 0.630 & 0.037 & -0.07 \\
GAE           & 0.056 & 0.363 & 0.327 & \textbf{0.585} \\
DynamicTriad  & 0.239 & 0.634 & 0.425 & 0.424 \\
Dynamic SDNE  & 0.281 & 0.134 & 0.085 & 0.502 \\
DynAERNN      & 0.072 & -0.098 & 0.027 & -0.396 \\
EvolveGCN     & 0.071 & 0.209 & 0.191 & 0.310 \\
MTSN          & 0.062 & 0.031 & 0.061 & 0.270 \\
TREND         & 0.041 & -0.458 & 0.051 & 0.233 \\
tdGraphEmbed  & 0.355 & \textbf{0.751} & \underline{0.508} & 0.485 \\
Graph2vec     & 0.332 & 0.647 & 0.136 & 0.424 \\
UGraphEmbed   & 0.304 & 0.363 & 0.038 & 0.252 \\
Sub2vec       & 0.106 & 0.534 & 0.115 & 0.413 \\
GraphERT      & \textbf{0.370} & \underline{0.703} & \textbf{0.521} & 0.455 \\
\midrule
EvoFormer     & \textbf{0.370} & 0.693 & \textbf{0.521} & \underline{0.537} \\
\bottomrule
\end{tabular}

\end{table}

\begin{table*}[t]
\centering
\footnotesize
\caption{Temporal segmentation performance on DBLP$_{1}$–DBLP$_{5}$ using ACC, NMI, and F1-macro.}
\label{tab:ex3}
\vspace{-1.8em} 
\setlength{\tabcolsep}{3pt} 
\renewcommand{\arraystretch}{0.85} 
\resizebox{\textwidth}{!}{%
\begin{tabular}{c|ccc|ccc|ccc|ccc|ccc}
\toprule
\multirow{2}{*}{Model} & \multicolumn{3}{c|}{DBLP$_{1}$} & \multicolumn{3}{c|}{DBLP$_{2}$} & \multicolumn{3}{c|}{DBLP$_{3}$} & \multicolumn{3}{c|}{DBLP$_{4}$} & \multicolumn{3}{c}{DBLP$_{5}$} \\ \cmidrule{2-16} 
                       & ACC  & NMI  & F1-macro & ACC  & NMI & F1-macro & ACC  & NMI  & F1-macro &ACC  & NMI  & F1-macro & ACC & NMI  & F1-macro \\ \midrule
tdGraphEmbed           & 0.56 & 0.653 & 0.525   & 0.566 & 0.68 & 0.548   & 0.50 & 0.679 & 0.467   & 0.538 & 0.657 & 0.467   & 0.631 & 0.732 & 0.606   \\ \midrule
node2vec               & 0.68 & 0.742 & 0.649   & \underline{0.733} & \underline{0.748} & \underline{0.713}   & 0.60 & 0.699 & 0.572  & 0.50  & 0.601 & 0.415   & 0.763 & 0.825 & 0.742   \\ \midrule
Graphert              & \underline{0.80} & \underline{0.776} & \underline{0.792}   & \textbf{0.866} & \textbf{0.809} & \textbf{0.865}   & \underline{0.766} & \underline{0.772}  & \underline{0.762}   & \underline{0.769} & \underline{0.734} & \underline{0.681}   & \underline{0.789} & \underline{0.785} & \underline{0.787}   \\ \midrule
EvoFormer              & \textbf{0.84} & \textbf{0.830} & \textbf{0.829}   & \textbf{0.866} & \textbf{0.809} & \textbf{0.865}   & \textbf{0.84} & \textbf{0.830} & \textbf{0.829}   & \textbf{0.866} & \textbf{0.809} & \textbf{0.865}   & \textbf{0.921} & \textbf{0.901} & \textbf{0.924}   \\ \bottomrule
\end{tabular}%
}
\end{table*}

\noindent{\bf Performance on Graph Similarity Ranking.} Table~\ref{tab:ex1} reports the results of graph similarity ranking across four benchmark datasets. EvoFormer achieves the best overall performance across the majority of metrics and datasets, consistently ranking at or near the top among all 19 baselines. 

On the ER dataset, EvoFormer achieves the highest scores across all four metrics, with relative improvements over GraphERT of 11.9\% (P@5), 9.4\% (P@10), 9.0\% (MRR), and 5.5\% (MAP@10). For R-GOT, which exhibits narrative-driven structural evolution, EvoFormer also outperforms all baselines, with gains up to 7.0\% (P@5), 4.1\% (P@10), 3.8\% (MRR), and 6.9\% (MAP@10) over the second-best method GraphERT. On the FB dataset, EvoFormer achieves improvements of 1.0\%, 0.9\%, and 1.4\% over the strongest baseline in P@5, P@10, and MAP@10, respectively. Although it does not achieve the best MRR, this may be related to the dataset's smooth temporal evolution, where short-term structural similarity plays a larger role. In such cases, models with stronger local sensitivity may perform better in early ranking positions. On the R-F1 dataset, EvoFormer improves P@5 and P@10 by 4.6\% and 7.7\% compared to the best baselines. Its performance in MRR and MAP@10 is slightly lower. This may be due to the dataset's stable structure over time, which reduces the effectiveness of phase-aware modeling designed for dynamic changes.

Figure~\ref{fig:heatmaps} shows the pairwise cosine similarity between graph embeddings over time. The FB dataset (a) displays a clear diagonal pattern, indicating smooth and consistent structural evolution. In contrast, the ER dataset (b) presents irregular and block-like patterns, with sharp drops in similarity across certain time intervals. This reflects periods of structural disruption, likely driven by organizational shifts or external events. These patterns support EvoFormer's advantage in modeling irregular and dynamic graph sequences.

\noindent{\bf  Performance on Temporal Anomaly Detection.} Table~\ref{tab:ex2} reports the results on R-GOT and R-F1 datasets using MRR and Spearman correlation. EvoFormer achieves an MRR of 0.370 on R-GOT, matching GraphERT and node2vec, and successfully identifies \emph{all} seven ground-truth anomalies. In Spearman correlation, EvoFormer ranks third (0.693), following tdGraphEmbed (0.751) and GraphERT (0.703). On the R-F1 dataset, EvoFormer also achieves the highest MRR (0.521), tied with GraphERT, and detects \emph{all} four annotated anomalies. In Spearman correlation, it ranks second (0.537), behind GAE (0.585). These results demonstrate that EvoFormer effectively identifies annotated anomalies while maintaining strong temporal ranking consistency, particularly on graphs with irregular structural changes.

\noindent{\bf Performance on Temporal Segmentation.} Table~\ref{tab:ex3} reports results on the dynamic segmentation task over five DBLP graph sequences. We compare EvoFormer against three strong baselines: tdGraphEmbed, node2vec, and GraphERT, using ACC, NMI, and F1-macro. EvoFormer achieves the highest performance on all three metrics across all datasets. Compared to the best-performing baseline GraphERT, EvoFormer improves segmentation quality notably. On DBLP1, EvoFormer achieves 0.84 (ACC), 0.83 (NMI), and 0.829 (F1-macro), representing relative gains of 5.0\%, 7.0\%, and 4.7\%. On DBLP2, both models reach identical scores. On DBLP3, EvoFormer improves over GraphERT by 9.7\% in ACC, 7.5\% in NMI, and 8.8\% in F1-macro. On DBLP4, the gains are larger, with 12.6\%, 10.2\%, and 27.0\% improvements across the three metrics. The most significant gains appear on DBLP5, where EvoFormer surpasses GraphERT by 16.7\%, 14.8\%, and 17.4\%. These results indicate that EvoFormer effectively captures temporal boundaries through its segmentation-aware self-attention.

\begin{table}[h]
\centering
\caption{Ablation results on ER and R-GOT datasets.}
\vspace{-1.5em}  
\label{tab:ex5}
\setlength{\tabcolsep}{2pt} 
\renewcommand{\arraystretch}{0.85} 
\small
\begin{tabular}{l|cccc|cccc}
\toprule
\multirow{2}{*}{Model}& \multicolumn{4}{c|}{ER} & \multicolumn{4}{c}{R-GOT} \\
\cmidrule{2-9}
&P@5 & P@10 & MRR & MAP@10 &P@5 & P@10 & MRR & MAP@10  \\
\midrule
Base &0.537  & 0.500 & 0.629 & 0.646  & 0.409 & 0.391 &0.471&0.589 \\
SAPE & 0.542 & 0.502 & 0.662 & 0.655 & 0.425 & 0.412 & 0.5 & 0.605 \\
SATA   & 0.547 & 0.519 & 0.636 & 0.651 & 0.461  &  0.419 &  0.509 & 0.617 \\
EvoFormer & 0.585& 0.547 & 0.703 & 0.673 & 0.445 & 0.434 & 0.497& 0.622 \\
\bottomrule
\end{tabular}
\end{table}

\noindent{\bf  Ablation Study. } Table~\ref{tab:ex5} presents the ablation results on ER and R-GOT, evaluating the individual contributions of Structure-Aware Positional Encoding (SAPE) and Segment-Aware Temporal Self-Attention (SATA). The base model removes both modules, using a Transformer with  timestamp and edge prediction. Adding SAPE alone yields consistent gains: on ER, the average improvement across four metrics is 1.99\%, while on R-GOT it reaches 4.54\%. This confirms SATM's effectiveness in mitigating structural bias and enhancing topology modeling. Incorporating only SATA leads to stronger gains on R-GOT (avg. +8.18\%) and a 1.89\% boost on ER, demonstrating its ability to capture abrupt transitions and non-uniform evolution. Combining both modules gives the best results. EvoFormer improves over the base model by 8.57\% on ER and 7.72\% on R-GOT (averaged across all metrics), highlighting the complementary benefits of structural encoding and temporal segmentation.

\begin{figure}[htbp]
    \centering
    \includegraphics[width=\linewidth]{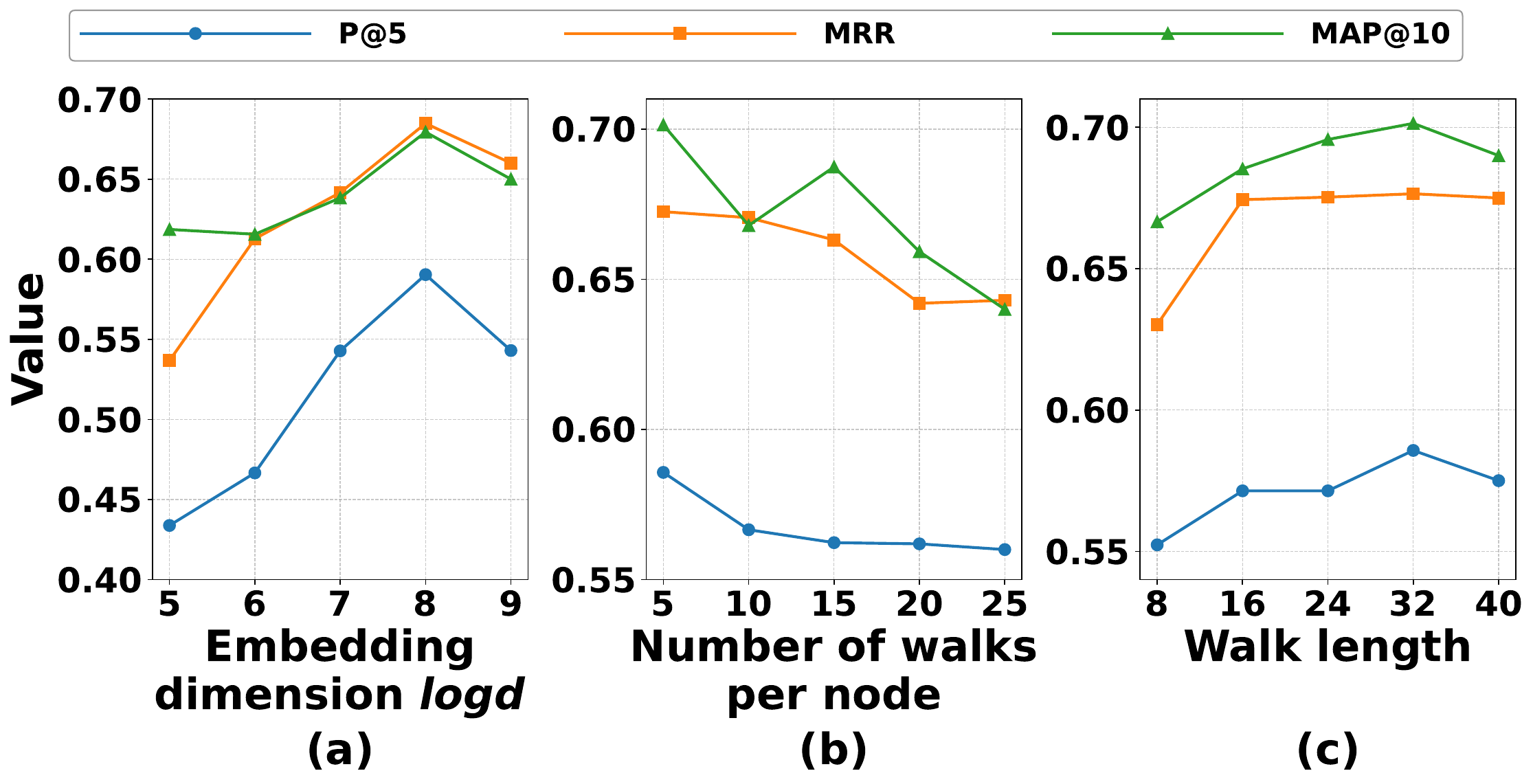}
    \vspace{-2.9em}
    \caption{Parameter sensitivity analysis of EvoFormer on graph similarity ranking.}
    \label{fig:sensitivity}
\end{figure}

\noindent{\bf Parameter Sensitivity Analysis.} We investigate the effect of three key hyperparameters on model performance: embedding dimension $d$, number of walks per node $W$, and walk length $L$. Results are shown in Figure~\ref{fig:sensitivity} using P@5, MRR, and MAP@10 as evaluation metrics. 
In Figure~\ref{fig:sensitivity} (a), increasing $\log_2 d$ from 5 to 8 leads to consistent performance gains across all metrics. The best results are observed at $d=256$, while a slight drop at $d=512$ indicates potential overfitting or redundancy at very high dimensions. Figure~\ref{fig:sensitivity} (b) shows that model performance is highest when $W=5$. Increasing $W$ beyond this point results in a decline in all metrics, suggesting that excessive sampling introduces noise or redundant walks, which may impair representation quality. As shown in Figure~\ref{fig:sensitivity} (c), longer walks improve performance up to $L=32$, which achieves the peak for all metrics. Beyond this, gains saturate or slightly decline, implying that very long walks may dilute useful local context.

\begin{figure}[htbp]
    \centering
    \includegraphics[scale=0.4]{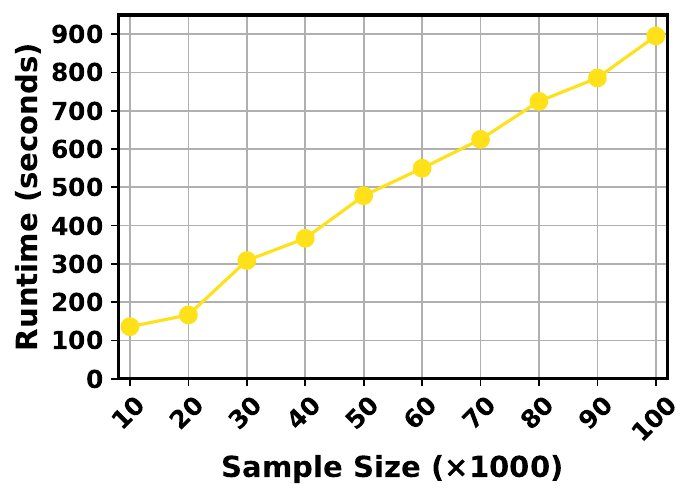}
    \vspace{-1.7em}
    \caption{Scalability analysis of EvoForme on different numbers of random walks}
    \label{fig:Scalability}
\end{figure}

\noindent{\bf Scalability.} We first report the per-batch execution time (batch size = 1) of each module across the FB, ER, R-GOT, and R-F1 datasets (Table~\ref{tab:ex6}). The Random Walk Timestamp Classification module is highly efficient, requiring only 0.16–0.17ms across all datasets, indicating that temporal label prediction is computationally negligible in our framework. The Structure‑Aware Transformer Module consistently takes around 12–13ms, showing stable performance with minor variations among datasets. In contrast, the Evolution‑Sensitive Temporal Module exhibits the highest computational cost, ranging from 18.06ms on FB to 31.6ms on R‑F1, which can be attributed to its modeling of long‑range temporal dependencies and more complex operations. To further examine scalability with respect to data volume, we varied the number of random walks from 10,000 to 100,000 and trained  R-F1 dataset for one epoch on a single NVIDIA RTX 4090 GPU. As shown in Figure~\ref{fig:Scalability}, the runtime increases approximately linearly with the sample size, from 150s (10k walks) to 900s (100k walks), confirming that EvoFormer scales efficiently to large networks.

\begin{table}[h]
\centering
\caption{Per-batch runtime (ms) of each module on different datasets.}
\vspace{-1.4em}
\label{tab:ex6}
\small
\setlength{\tabcolsep}{2pt} 
\renewcommand{\arraystretch}{0.9} 
\begin{tabular}{lcccc}
\toprule
\multirow{2}{*}{Component} & \multicolumn{4}{c}{Execution Time / Batch Size = 1} \\
\cmidrule(lr){2-5}
 & FB & ER  & R-GOT& R-F1 \\
\midrule
 Structure-Aware Transformer & 12.76 & 12.09 & 12.04 & 13.47 \\
Timestamp Classification &  0.17 & 0.16 & 0.16 & 0.17 \\
Evolution-Sensitive Temporal Module  &  18.06 &  19.8 & 28.9 & 31.6 \\
\bottomrule
\end{tabular}
\end{table}

\section{Conclusion}

In this paper, we introduced EvoFormer, a novel Transformer-based approach designed to effectively address two critical limitations in dynamic graph embedding: Structural Visit Bias and Abrupt Evolution Blindness. Our framework integrates a Structure-Aware Transformer Module, employing structure-informed positional encoding to mitigate redundancy and noise in node representations, and an Evolution-Sensitive Temporal Module, featuring timestamp classification, graph-level segmentation, segment-aware temporal attention, and edge prediction, enabling accurate detection of sudden structural transitions. Future work includes extending EvoFormer to heterogeneous and continuous-time dynamic graphs, and exploring its applications in other domains with evolving structures, such as bioinformatics and cybersecurity.

\nocite{*}

\begin{acks}
This work was supported by the National Natural Science Foundation of China (62402368); the Outstanding Youth Science Foundation of Shaanxi Province (2025JC-JCQN-083); the Key Research and Development Program of Shaanxi Province (2025CY-YBXM-047); the Young Talent Fund of Xi'an Association for Science and Technology (0959202513020); the Natural Science Basic Research Program of Shaanxi Province (2024JC-YBQN-0715); and the Fundamental Research Funds for the Central Universities of China (XJSJ24015, QTZX23084).
\end{acks}

\section*{GenAI Usage Disclosure}

The authors confirm that no generative AI tools were used at any stage of the research process, including manuscript writing and code.

\bibliographystyle{ACM-Reference-Format}
\balance
\bibliography{reference}

\end{document}